\documentclass{article}

\PassOptionsToPackage{square,numbers}{natbib}

\usepackage[preprint]{neurips_2025}

\usepackage[utf8]{inputenc} 
\usepackage[T1]{fontenc}    
\usepackage{hyperref}       
\usepackage{url}            
\usepackage{booktabs}       
\usepackage{amsfonts}       
\usepackage{nicefrac}       
\usepackage{amsmath}
\usepackage{wrapfig}

\usepackage{microtype}      
\usepackage{graphicx}
\usepackage[table,xcdraw]{xcolor}
\usepackage{enumitem}

\newcommand\bitsmall{\fontsize{7.8}{9}\selectfont}

\title{Bridging Annotation Gaps: Transferring Labels to Align Object Detection Datasets}

\author{Mikhail Kennerley\\
  National University of Singapore\\
  I2R, A*STAR\\
  \texttt{mikhailk@u.nus.edu}\\
  \And
  Angelica Aviles-Rivero\\
  Tsinghua Univeristy\\
  \texttt{aviles-rivero@tsinghua.edu.cn}\\
  \AND
  Carola-Bibiane Schönlieb\\
  University of Cambridge\\
  \texttt{c.b.schoenlieb@damtp.cam.ac.uk}\\
  \And
  Robby T. Tan\\
  National University of Singapore\\
  ASUS Intelligent Cloud Services\\
  \texttt{robby.tan@nus.edu.sg} \\
}

\begin{document}

\maketitle

\begin{abstract}

Combining multiple object detection datasets offers a path to improved generalisation but is hindered by inconsistencies in class semantics and bounding box annotations.  
Some methods to address this assume shared label taxonomies and address only spatial inconsistencies; others require manual relabelling, or produce a unified label space, which may be unsuitable when a fixed target label space is required.  
We propose \textit{Label-Aligned Transfer (LAT)}, a label transfer framework that systematically projects annotations from diverse source datasets into the label space of a target dataset.  
LAT begins by training dataset-specific detectors to generate pseudo-labels, which are then combined with ground-truth annotations via a \textit{Privileged Proposal Generator (PPG)} that replaces the region proposal network in two-stage detectors.  
To further refine region features, a \textit{Semantic Feature Fusion (SFF)} module injects class-aware context and features from overlapping proposals using a confidence-weighted attention mechanism.  
This pipeline preserves dataset-specific annotation granularity while enabling many-to-one label space transfer across heterogeneous datasets, resulting in a semantically and spatially aligned representation suitable for training a downstream detector. 
LAT thus jointly addresses both class-level misalignments and bounding box inconsistencies without relying on shared label spaces or manual annotations.
Across multiple benchmarks, LAT demonstrates consistent improvements in target-domain detection performance, achieving gains of up to +4.8AP over semi-supervised baselines.

\end{abstract}

\section{Introduction}


\begin{figure}[t]
  \centering
  \includegraphics[width=0.75\linewidth]{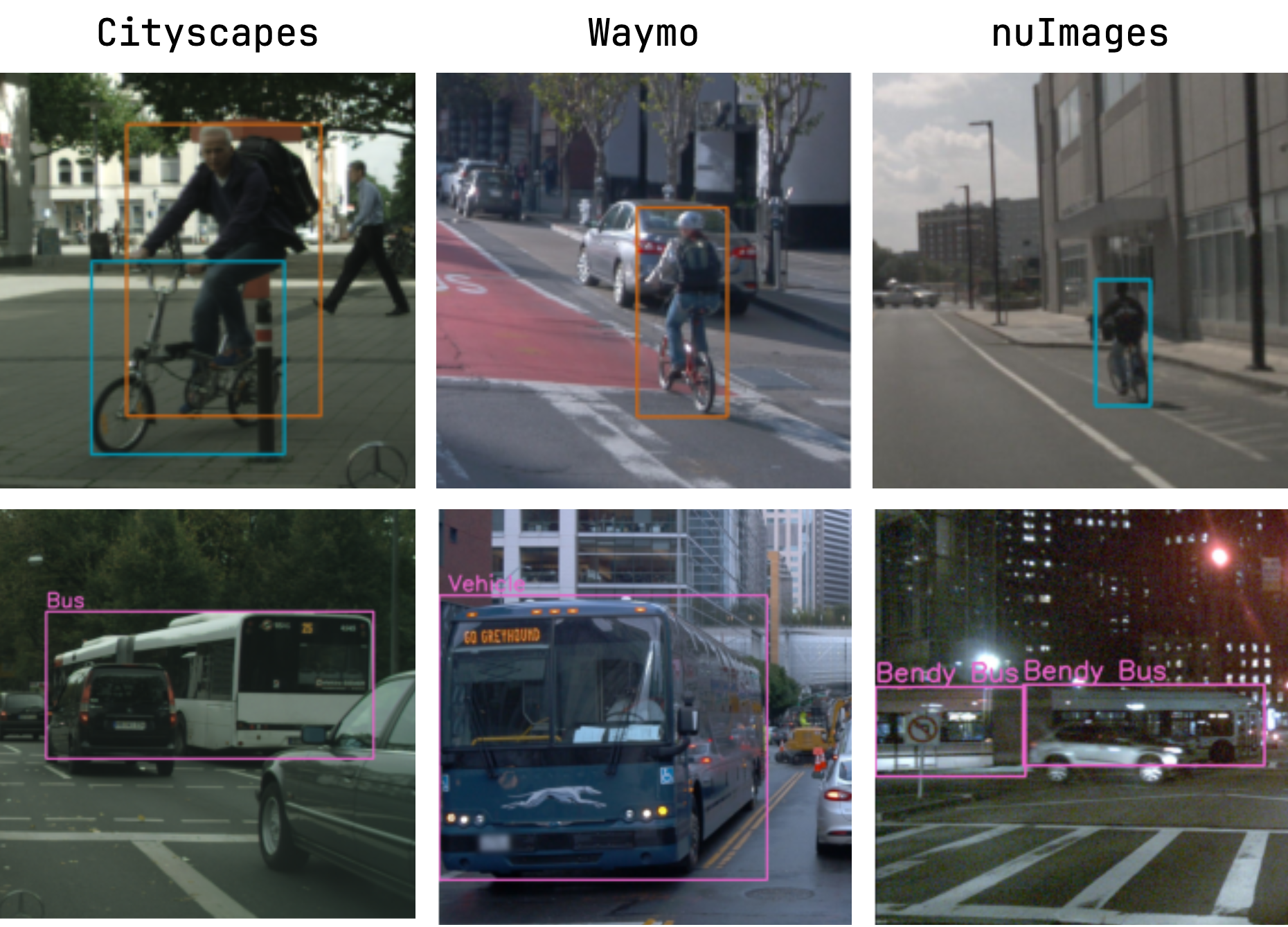}
  \caption{
Annotation discrepancies between three road-based object detection datasets: Cityscapes, Waymo, and nuImages. The top row highlights differences in cyclist-related annotations, Waymo and nuImages treat the cyclist and bicycle as a single entity but assign different class labels, whereas Cityscapes annotates them separately. The bottom row illustrates differences in label granularity, with nuImages exhibiting the most fine-grained annotations and Waymo the least.
}
\end{figure}

Combining multiple datasets is an increasingly practical strategy for improving object detection models, especially in domains where training data is limited or expensive to obtain. 
However, merging datasets naively with differing label spaces introduces annotation conflicts in class semantics, labelling granularity, background definitions, or bounding box styles \cite{Chen_2023_CVPR,Wang_2019_CVPR}. 
These mismatches lead to reduced performance on downstream tasks, particularly when the target dataset’s label space is task-specific or domain-sensitive \cite{liao2024transferring}. 
Manual relabelling is time-consuming and, when class semantics diverge significantly, may be equivalent to annotating from scratch.


Recent model-centric efforts mitigate dataset inconsistencies by learning shared taxonomies or semantic spaces.
They achieve this by using vision-language alignment \cite{radford2021learning, ilharco_gabriel_2021_5143773,shi2024plaindet, Chen_2023_CVPR,zhou2023lmseg,Meng_2023_CVPR} or graph-based structures \cite{unircnn, ma2024automated} to unify labels during multi-dataset training.
However, these approaches primarily optimise for average-case generalisation, where a unified label space is needed.
While this unified label space is beneficial for out-of-domain classification, they are not designed to prioritise fidelity to any particular target label spaces.
This limits their practical usage in settings where high precision on a specific dataset is essential.

Data-centric approaches \cite{liao2024transferring,lambert2020mseg}, which transfer the annotations of datasets from their original label space to a target label space, addresses the issue of requiring a specific annotation style. 
However, these methods are limited in that they require manual relabelling \cite{lambert2020mseg}, only transfer one dataset to another at a time, or only account for bounding box styles \cite{liao2024transferring} rather than both bounding box styles and class semantics in unison.
This severely limits the usefulness of these methods in realistic settings where multiple datasets would be needed as well as alignment of both class semantics and bounding box styles to the target space.


In our work, we propose a data-centric approach to multi-dataset object detection.
Rather than enforcing a unified label space across datasets, we transfer annotations from source datasets into the label space of a given target. 
This allows practitioners to improve detection performance on a small, use-case-specific dataset by leveraging external data \cite{liao2024transferring}, without altering their annotation conventions or model architecture.
Crucially, our approach addresses both semantic and bounding box misalignments, enabling targeted integration of large-scale datasets while preserving semantic alignment with the user’s intended label space.
%


Our approach begins by training a separate object detector for each dataset in a given set.
This allows each model to specialise in its native annotation style, or label space.
Once trained, these dataset-specific models are used to generate cross-dataset pseudo-labels.
Where for each dataset, predictions are made in the label spaces of the other N-1 datasets, where N is the total number of datasets in the set.
This results in a total of N(N-1) pseudo-label mappings, each providing an implicit alignment between a pair of datasets.
We leverage the ground-truth annotations as supervision and treat the projected pseudo-labels as a bridge to infer label correspondences.
This multi-source supervision allows the model to learn semantic and spatial correspondences between datasets.

Unlike conventional pseudo-labelling \cite{pseudolabel} or semi-supervised methods \cite{liu2021unbiased,kennerley2023tpcnet,kennerley2024cat,li2022cross,hoffman2018cycada}, our method preserves the information content of source labels while aligning them to target conventions. 
This is achieved through our Label-Aligned Transfer (LAT) framework, which combines projected pseudo-labels and ground-truth annotations to learn semantic and spatial correspondences across datasets.
Our Privileged Proposal Generator (PPG) replaces the conventional region proposal network by injecting fused pseudo- and ground-truth proposals into the detection pipeline.
Additionally, our Semantic Feature Fusion (SFF) module refines region features using class-aware and overlap-sensitive attention. 
We validate our framework across multiple object detection benchmarks and show that it consistently improves performance, outperforming baseline semi-supervised learning strategies. 
By shifting the emphasis from unified model architectures to annotation-aware data transformation, our approach enables scalable and flexible integration of heterogeneous datasets in real-world object detection systems.
%
%
%
In summary, our contributions are as follows.
\begin{itemize}
      \item We introduce \textbf{Label-Aligned Transfer (LAT)}, a data-centric framework that systematically transfers both semantic class labels and spatial bounding boxes from multiple source datasets into a target label space, without requiring manual taxonomy merging.
    \item We design two novel components to support robust cross-dataset label transfer: the \textbf{Privileged Proposal Generator (PPG)}, which injects ground-truth and cross-space pseudo-labels as region proposals, and the \textbf{Semantic Feature Fusion (SFF)} module, which uses class-aware attention to model inter-dataset relationships and inject privileged semantic context into the detection pipeline.
    \item We evaluate LAT across diverse object detection settings, datasets with mismatched label granularity and with large-scale size disparity, and demonstrate consistent improvements over state-of-the-art baselines, including a \textbf{+4.2AP gain} on high-low class alignment and \textbf{+4.8AP} on small-large dataset transfer.
\end{itemize}

\section{Related Work}

\paragraph{Dataset Alignment}

Integrating datasets for object detection presents challenges beyond visual domain shifts, including semantic misalignment and inconsistent annotation protocols.  
Early work in domain adaptation focused on aligning image distributions via model-level adjustments such as Maximum Mean Discrepancy (MMD)~\cite{yan2017mind}, domain-adversarial training~\cite{hoffman2018cycada}, and self-training~\cite{kennerley2024cat, kennerley2023tpcnet, li2022cross}.  
Other approaches adopt a data-centric view, employing image translation to harmonise low-level appearance features across domains~\cite{zheng_2020_ECCV,2023iccv_PASTA}.  
However, these methods primarily address distributional variance and do not resolve inconsistencies in annotation semantics or structure.

Annotation mismatches have been more extensively studied in classification~\cite{pmlr-v97-recht19a, beyer2020imagenet, yun2021revisiting} and semantic segmentation~\cite{autotax, Bevandic_2022_WACV, Rottmann_2023_WACV, ma2024automated}, while object detection remains underexplored.  
Our framework addresses this gap by jointly correcting semantic and spatial inconsistencies via direct label transfer into a designated target label space, without requiring manual taxonomies or repeated re-labelling.

\paragraph{Multi-Dataset Object Detection}

Multi-dataset training is commonly used to improve robustness and expand object category coverage~\cite{Chen_2023_CVPR, ma2024automated, Meng_2023_CVPR}.  
Approaches can be grouped into three broad categories: (1) partitioned detectors with dataset-specific heads~\cite{Zhou_2022_CVPR, shi2024plaindet}, (2) unified detectors trained on merged label spaces~\cite{Wang_2019_CVPR, Chen_2023_CVPR}, and (3) hybrid models that incorporate pseudo-labelling across datasets~\cite{liao2024transferring}.  
To unify label semantics, early work relied on manual class mapping and taxonomy construction~\cite{lambert2020mseg}, while more recent approaches use vision-language models~\cite{radford2021learning, ilharco_gabriel_2021_5143773} to build shared embedding spaces, enabling prompt-based alignment across datasets~\cite{shi2024plaindet, Chen_2023_CVPR, zhou2023lmseg, Meng_2023_CVPR}.  
While effective at harmonising class names, these methods often overlook differences in annotation coverage or bounding box conventions, and typically generate generalised label spaces rather than adapting to a task-specific target ontology.  

In contrast, LAT transfers annotations directly into a fixed target label space, eliminating the need for dataset-specific heads or handcrafted taxonomies.  
Unlike previous methods that aim to optimise average performance across datasets, our approach is designed to maximise target-domain performance critical for real-world deployments with specific annotation requirements.

\section{Method}



We propose \textbf{Label-Aligned Transfer (LAT)} to address the challenges associated with combining object detection datasets that exhibit differing label spaces.  
Our \textit{data-centric framework} transfers annotations from multiple source datasets into the label space of a designated target dataset, without requiring unified taxonomies.  
This is accomplished through \textit{collaborative pseudo-labeling} across heterogeneous label spaces, enabling the alignment of annotations at both the semantic and spatial levels.  
In doing so, LAT facilitates the transfer of not only class semantics but also bounding box conventions, effectively bridging inter-dataset discrepancies.  

\subsection{Preliminaries}
We begin by defining the formal problem setup and describing the steps involved in generating initial pseudo-labels across datasets with divergent label spaces.

\subsubsection{Problem formulation}

Given $N$ datasets with respective label spaces $\{L_1, L_2, \ldots, L_N\}$, our goal is to transfer annotations from all datasets into the label space of a target dataset, formalised as $\{L_1, L_2, \ldots, L_{N-1}\} \rightarrow L_N$.  
We assume that the datasets differ not only in bounding box conventions but also in class semantics.  
This substantially increases the difficulty of label transfer compared to prior approaches, which often assume a shared or compatible class labels across datasets~\cite{liao2024transferring}.

Our label transfer model operates on an image, pseudo-label, ground-truth triplet, where the pseudo-labels represent the set $\{PL^n_1, PL^n_{n-1}, PL^n_{n+1}, \ldots, PL^n_N\}$.  
The model outputs a refined set of pseudo-labels aligned with a selected target label space.  
Ultimately, our objective is to generate target-aligned annotations for all datasets, enabling downstream training of object detectors within a consistent label space.

The primary challenge lies in the absence of paired supervision: we do not observe ground-truth annotations in both the source and target label spaces for any single image.  
Moreover, datasets may exhibit class label sparsity, semantic overlap, or divergent naming conventions.  
To address these issues, LAT performs a many-to-one label space transfer, allowing pseudo-labels from different source domains to reinforce one another in a collaborative, ensemble-like training process.

\begin{figure}[t]

  \centering
  \includegraphics[width=0.95\linewidth]{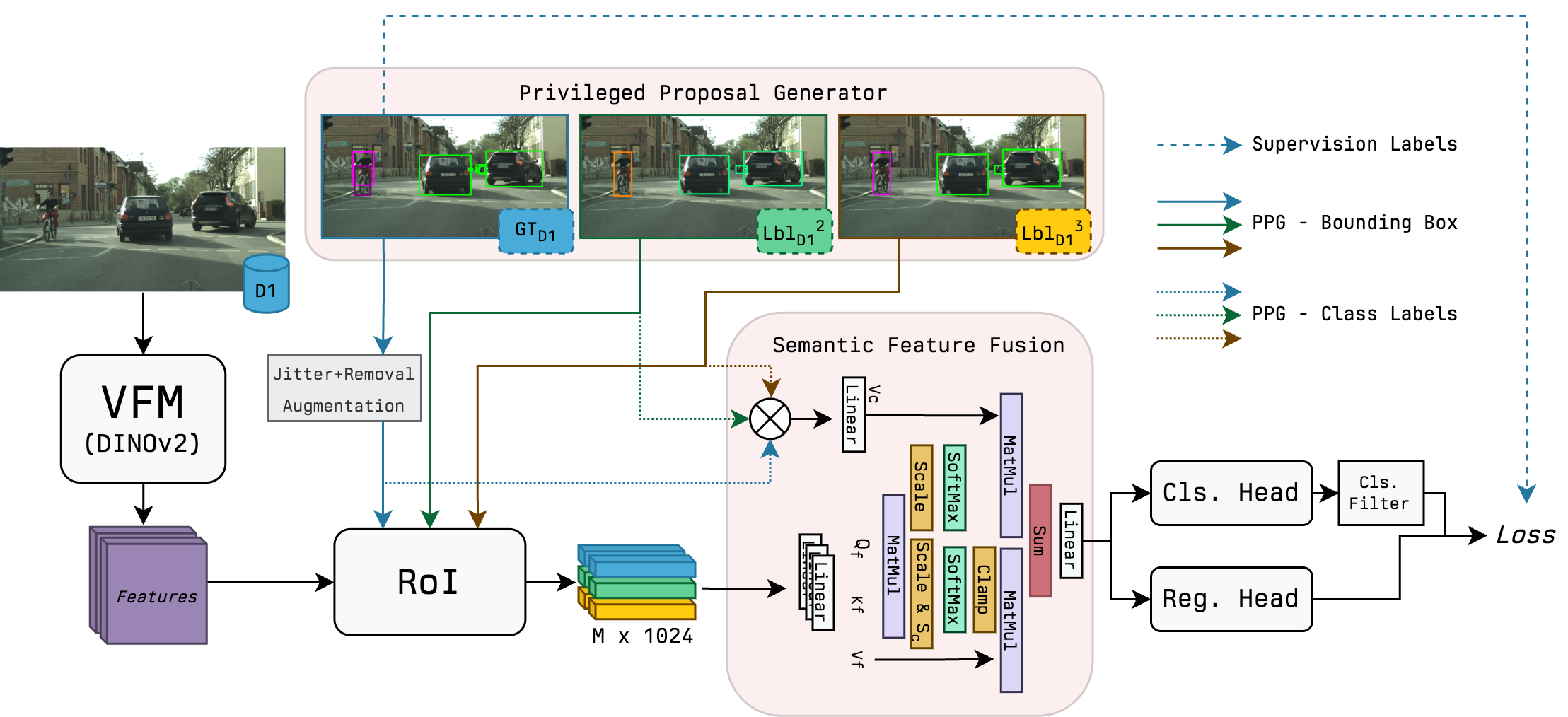}

  \caption{Overview of the LAT architecture. Dataset-specific pseudo-labels and ground-truth annotations are combined via the Privileged Proposal Generator (PPG), which replaces the region proposal network. A frozen Vision Foundation Model (VFM) extracts shared image features. The Semantic Feature Fusion (SFF) module then refines region features by injecting class-aware information using attention over overlapping proposals. We filter the classification output to compute loss on only the current datasets label space.}
\label{fig:arch}
\end{figure}

\subsubsection{Generating initial pseudo-labels} \label{sec:generatelabels}

We begin with a set of datasets $\{D_1, D_2, \ldots, D_N\}$, where $N$ denotes the total number of datasets.  
For each dataset $D_n$, we train a corresponding object detector $M_n$, which is optimised on its native label space $L_n$.  
Using these trained models, we generate pseudo-labels for each dataset under every other dataset's label space.  
In effect, each dataset yields $N-1$ sets of pseudo-labels, corresponding to the annotation formats of the remaining datasets.  
These pseudo-labels are accompanied by their associated classification confidence scores, which are retained for downstream use.  
To improve reliability, we apply non-maximum suppression and score thresholding to filter out low-confidence predictions.

These pseudo-labels serve as the initial cross-space predictions for each dataset.  
However, they are generated independently by each model and do not incorporate additional contextual signals that could enhance their accuracy.  
We refer to such contextual signals as \textit{privileged information}, which is typically unavailable in standard supervised training settings.  
In our framework, this privileged information includes the ground-truth labels (classes and bounding boxes) as well as pseudo-labels from other label spaces.  




\subsection{Label-Aligned Transfer}

Our Label-Aligned Transfer (LAT) model extends the standard two-stage object detector to incorporate privileged information during both training and inference.  
A typical two-stage detector \cite{fasterrcnn} consists of three main components: a feature extractor $f_{img}$, a region proposal network (RPN), and a region-of-interest (RoI) pooling layer.  
The RPN generates class-agnostic bounding box proposals from the feature maps, while the RoI layer extracts fixed-size region features that are passed to the classification and regression heads.

In LAT, we replace the conventional feature extractor with a frozen Vision Foundation Model (VFM), such as DINOv2~\cite{oquab2024dinov}.  
We further substitute the RPN with our \textbf{Privileged Proposal Generator (PPG)}, which provides high-quality region proposals derived from both ground-truth and pseudo-label sources.  
These proposals are passed to the RoI layer and also serve as input to the \textbf{Semantic Feature Fusion (SFF)} module, which refines region features using class-aware attention.  

\subsubsection{Privileged Proposal Generator (PPG)} \label{sec:ppg}

Our Privileged Proposal Generator (PPG) consists of pseudo-labels generated as described in Section~\ref{sec:generatelabels}, alongside the ground-truth labels for each image in the training batch.  
We apply light augmentations to the ground-truth labels, such as random jittering and selective removal of bounding boxes.  
These augmented labels are then used by the RoI layer to crop region features from the shared feature map.  
Since these labels are derived from multiple label spaces, they often contain overlapping objects across datasets, regardless of naming convention.  
For example, the concept of a \textit{car} may appear across datasets but be labeled as \textit{vehicle} in Waymo and \textit{car} in Cityscapes.  
Such overlaps provide a rich supervisory signal and are critical to the effectiveness of our Semantic Feature Fusion (SFF) module.

In addition to supplying bounding box proposals to the RoI layer, PPG also outputs the associated class labels for each region to the SFF module.  
To maintain label discreteness, we concatenate the label sets from all datasets instead of merging classes with identical names.  
This ensures that semantically divergent classes, despite having the same label name, are not erroneously unified.  
We elaborate on the usage of these class labels and their role in fusion in Section~\ref{sec:sff}.

%

%
%
%
%
%
%

\subsubsection{Semantic Feature Fusion (SFF)} \label{sec:sff}

\begin{figure}[t!]
\centering
\begin{minipage}{.488\textwidth}
  \centering
  \includegraphics[width=.95\linewidth]{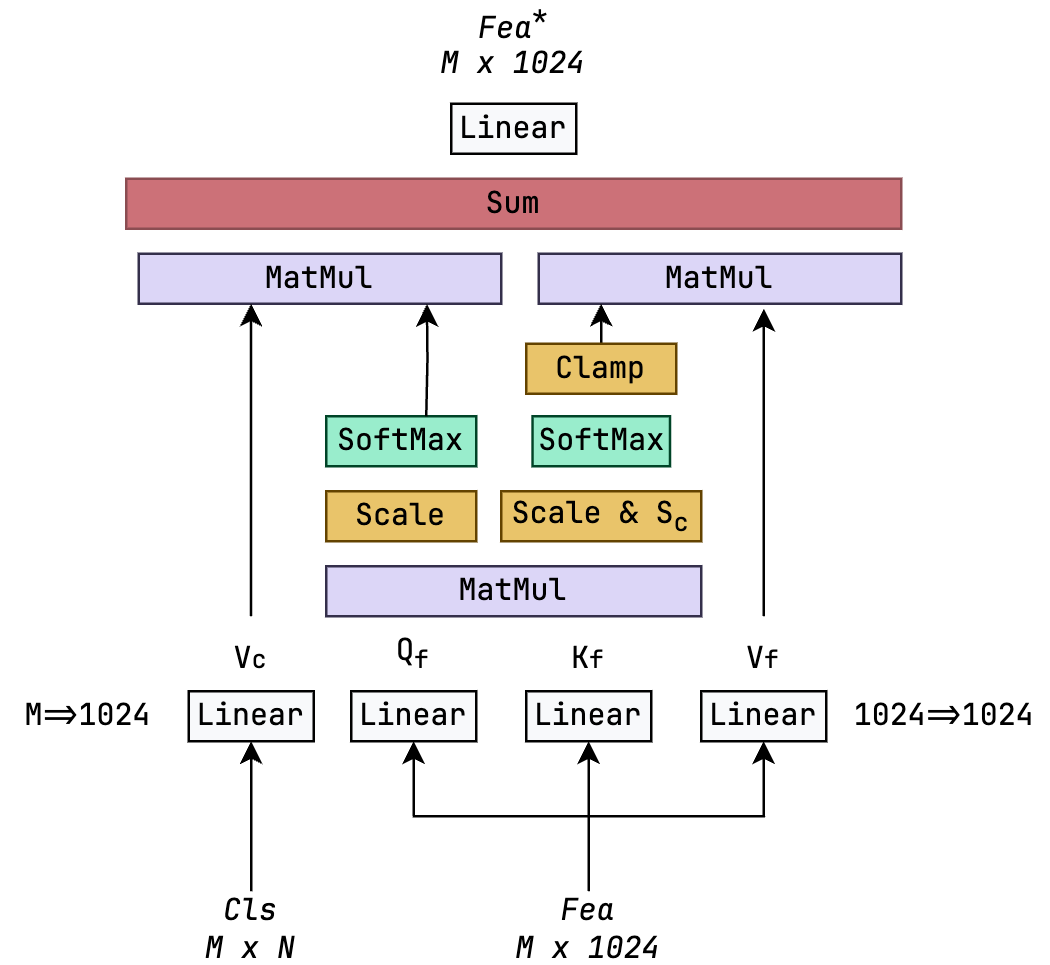}
  \caption{Our Semantic Feature Fusion (SFF) module leverages an attention mechanism to fuse and inject information from overlapping proposals. This enhances the capacity of the downstream classification and regression heads to model relationships between classes originating from distinct label spaces.}
  \label{fig:sff}
\end{minipage}%
\hspace{0.02\textwidth}%
\begin{minipage}{.488\textwidth}
  \centering
  \includegraphics[width=.95\linewidth]{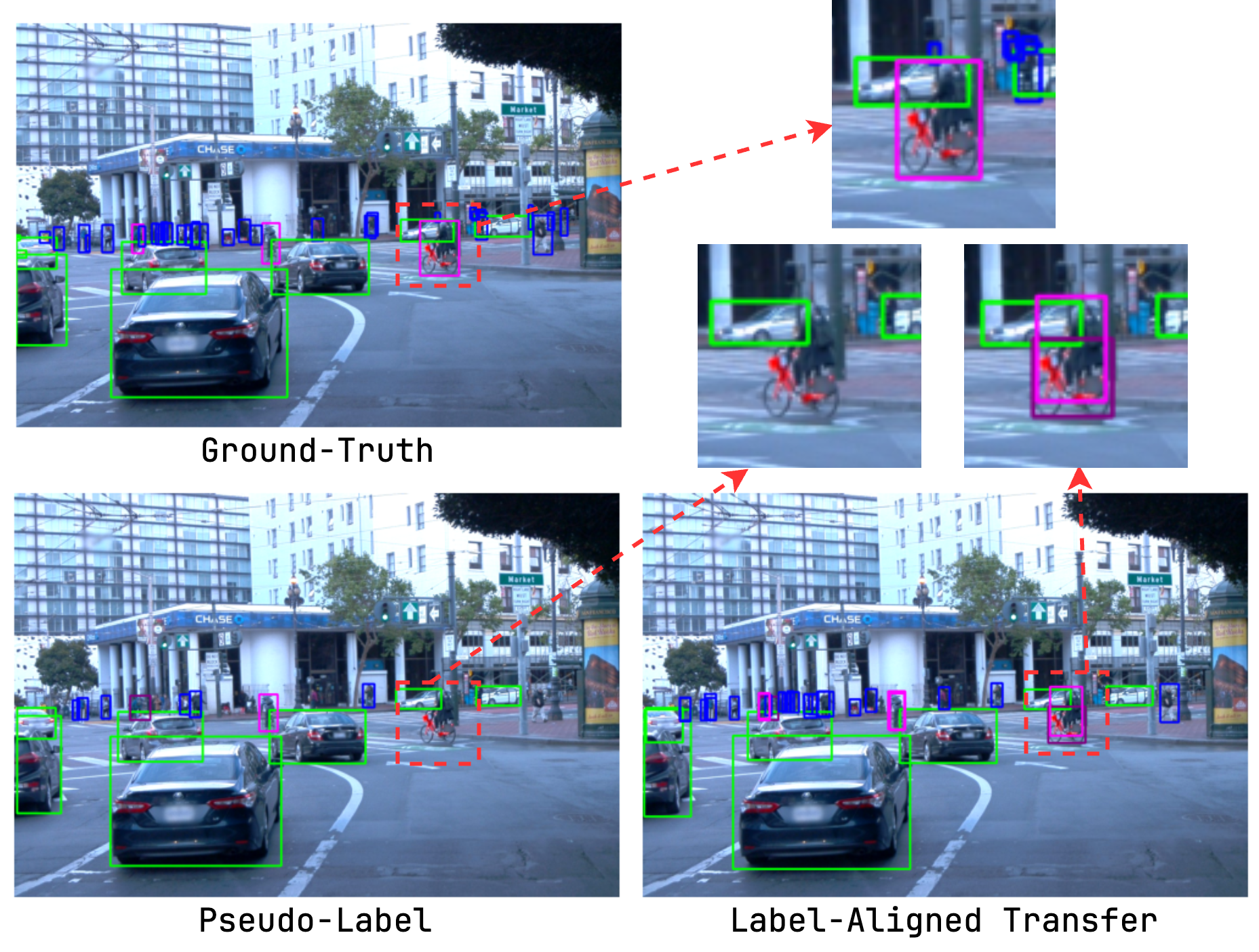}
  \caption{Comparison of labels on a Waymo image. Ground-Truth labels are the original Waymo labels. Pseudo-labels and Label-Aligned Transfer are in the Cityscapes label space. Label-Aligned Transfer accurately labels the cyclist and bicycle into their respective classes and boxes. Privileged ground-truth information also results in more accurate detections of smaller objects missing in the pseudo-labels. }
  \label{fig:test2}
\end{minipage}
\end{figure}

To improve cross-dataset feature consistency, we introduce the Semantic Feature Fusion (SFF) module (Figure~\ref{fig:sff}). 
SFF enhances region features by attending over overlapping proposals and injecting class-aware information from both pseudo-labels and ground-truth annotations. 
This allows the model to learn semantic relationships between related but differently labelled classes.

Let \( A \in \mathbb{R}^{M \times M} \) denote the attention matrix computed using scaled dot-product attention over RoI features:
\[
A = \frac{QK^T}{\sqrt{d}},
\]
where \( Q, K \in \mathbb{R}^{M \times d} \) are learned projections of the RoI features. Let \( V_c \in \mathbb{R}^{M \times d} \) be the value matrix derived from classification scores, and \( V_r \in \mathbb{R}^{M \times d} \) be the value matrix derived from region features. We define a confidence vector \( S_c \in \mathbb{R}^{M} \), where each entry is set to 1 for ground-truth proposals, or \( \max(C_m) \) for pseudo-label proposals, where \( C_m \) is the classification score vector of the \( m \)-th proposal.

To enhance the fusion of consistent semantics across datasets, we apply a row-wise clamping mechanism to the similarity matrix. Specifically, we scale each row of \( A \) such that its maximum value is equal to a threshold \( T = 1/\sqrt{N} \), where \( N \) is the number of datasets. This emphasises high-confidence, overlapping proposals while reducing noise from spurious predictions. In parallel, we compute an unclamped version of \( A \) for use with the classification value matrix.

The final fused feature representation is computed as:
\[
SA = \text{softmax}(A) V_c + \text{clamp}\left( \text{softmax}(S_c \circ A) \right) V_r,
\]

where \( \circ \) denotes element-wise multiplication. Softmax is applied row-wise, and clamping ensures that each row’s maximum value does not exceed \( T \). This fused output provides the downstream classifier with enriched semantic and visual features, improving cross-dataset generalisation.


%
%

SFF enables the classification and regression heads of LAT to leverage enriched visual features and semantic context from overlapping label spaces.  
During training, we mask the classification logits to include only the classes present in the current batch, ensuring that intra-dataset supervision remains dominant while still benefiting from inter-dataset relationships.  
At inference, logits are masked to the designated target label space.  
By learning cross-dataset semantic correspondences, LAT can generate robust predictions even for datasets not originally labelled under the target space.  
We demonstrate the effectiveness of this approach in Section~\ref{sec:experiments}.

\section{Experiments}\label{sec:experiments}
We conduct extensive experiments to evaluate the effectiveness of our proposed Label-Aligned Transfer (LAT) framework in realistic multi-dataset object detection scenarios. 
Our evaluations focus on two primary challenges: (1) semantic inconsistencies due to divergent class taxonomies across datasets, and (2) performance degradation in small datasets when augmented with larger ones.
These benchmarks simulate practical conditions where direct dataset merging would be ineffective or harmful. 

\subsection{Benchmark Datasets Description}

\begin{figure}[t]
  \centering
  \includegraphics[width=0.7\linewidth]{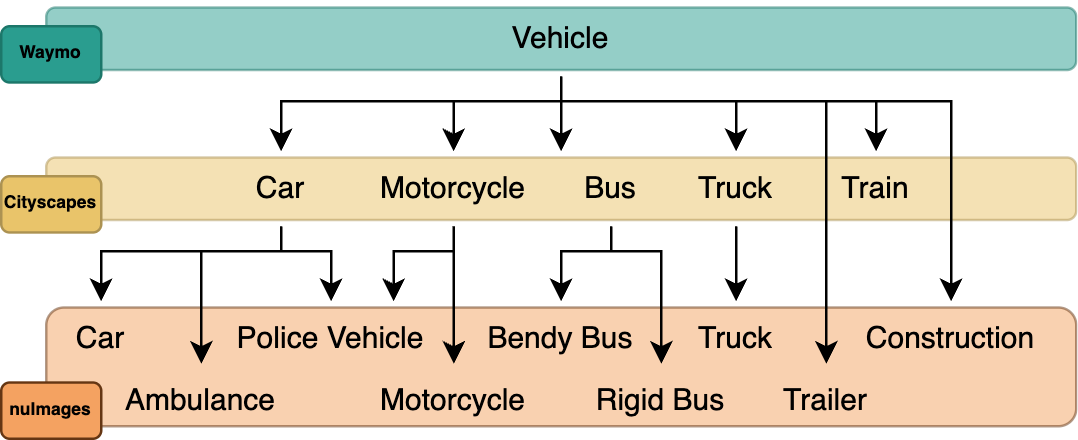}

  \caption{We visualise the labelling hierarchy of vehicle classes in the  Waymo, Cityscapes, and nuImages datasets which are ordered from least to most discrete.} \label{fig:tree}
\end{figure}

\paragraph{Cityscapes $\leftrightarrow$ nuImages $\leftrightarrow$ Waymo.}
This benchmark targets the challenge of class divergence across datasets.
Cityscapes~\cite{Cordts2016Cityscapes}, nuImages~\cite{nuscenes2019}, and Waymo~\cite{Sun_2020_CVPR} contain 8, 24, and 3 annotated classes, respectively, with varying levels of granularity.
An example of vehicle class hierarchy is illustrated in Figure~\ref{fig:tree}.
In this case, the \textit{vehicle} class in Waymo subsumes five vehicle-related classes in Cityscapes and nine in nuImages.
Conversely, several Cityscapes classes act as superclasses for their counterparts in nuImages.
One such ambiguity arises in nuImages, where the class \textit{police vehicle} encompasses instances that would be separately labeled as \textit{car} or \textit{motorcycle} in Cityscapes, illustrating inconsistencies in label boundaries across datasets.
To isolate the effects of class variability, we subsample 3,000 images from each dataset.%

%
%

%
%
%
%

\paragraph{Cityscapes $\leftrightarrow$ ACDC $\leftrightarrow$ BDD100K $\leftrightarrow$ SHIFT.}
This benchmark evaluates performance under a small-versus-large dataset setting.  
Such a scenario reflects real-world conditions, where annotating hundreds of thousands of images may be prohibitively costly or time-consuming.  
Cityscapes~\cite{Cordts2016Cityscapes} and ACDC~\cite{SDV21} represent small-scale datasets, each comprising of \textit{2,965} and \textit{1,571} samples, respectively.  
In contrast, BDD100K~\cite{Yu_2020_CVPR} and SHIFT~\cite{shift2022} are large-scale datasets, each containing \textit{69,852} and \textit{141,052} images.  
These four datasets were chosen not only for their variation in dataset size, but also for their consistency in class semantics.  
This ensures that the benchmark isolates the effect of sample size on performance, particularly with respect to bounding box variation.  

%
%
%
%
%
%

\subsection{Experimental Set-up}
We implement LAT using the Faster R-CNN~\cite{fasterrcnn} framework built on Detectron2~\cite{wu2019detectron2}.  
DINOv2~\cite{oquab2024dinov} is employed as a frozen feature extractor with pre-trained weights.  
In our PPG module, random jittering and ground-truth label removal are applied at rates of 0.5 and 0.05, respectively.  
LAT is trained for 30,000 iterations using a learning rate of 0.2 and a batch size of 4.  

For downstream training, we use a Faster R-CNN model with a modified weighted cross-entropy loss, where the weight is derived from the confidence score of the pseudo-label.  
This model, as well as the initial pseudo-label generation model, is trained for 50,000 iterations with a fixed learning rate of 0.2 and a batch size of 16.  
An exponential moving average (EMA) of the model weights is maintained for the final downstream detector for evaluation.  
All models are trained using two NVIDIA RTX 3090 GPUs.  

\paragraph{Baselines.}
We compare LAT against several methods, including a baseline without label transfer and two semi-supervised approaches: student-teacher supervision and pseudo-labelling.
    \setlist{nolistsep}

\begin{itemize}[noitemsep]
    \item \textbf{Baseline:} A model trained solely on the target dataset using its native label space.
    \item \textbf{Student-Teacher}~\cite{liu2021unbiased}: Trained with full supervision on the target dataset and semi-supervised loss on unlabelled samples from other datasets.
    \item \textbf{Pseudo-Label}~\cite{pseudolabel}: A model is first trained on the target dataset and then used to generate pseudo-labels on other datasets. These labels are filtered using non-maximum suppression and confidence thresholding before being used in continued training.
\end{itemize}
All baseline models adopt exponential moving average (EMA) updates for fair comparison.

\subsection{Results}
\begin{table}[t!]
  \caption{\textbf{Downstream AP} for detectors trained with and without label transfer in the \textbf{high–low class setting}. LAT outperforms all baselines; Cityscapes benefits most despite having moderate class count.
}
  
  \label{tab:classbench}
  \centering
  \begin{tabular}{lllll}
    \toprule
    & & \multicolumn{3}{c}{\textsc{Dataset}}                   \\
    \cmidrule(r){3-5}
    \textsc{Model}  & \textsc{Method}    & Cityscapes     & nuImages & Waymo  \\
    \midrule
    Baseline & - & 55.2  & 39.2 & 44.6     \\
    Student-Teacher & Semi-Supervised & 55.1  & 40.1 & 44.2     \\
    Pseudo-Label & Label Transfer & 56.9  & 40.6 & 45.6     \\
    LAT & Label Transfer & \cellcolor[HTML]{FFEBDB}\textbf{59.6} & \cellcolor[HTML]{FFEBDB}\textbf{41.5}  & \cellcolor[HTML]{FFEBDB}\textbf{47.9}     \\
    \bottomrule
  \end{tabular}
\end{table}

\begin{table}[t!]
  \caption{\textbf{Downstream AP} with and without label transfer in the \textbf{small–large dataset setting}. ACDC benefits most due to limited data; larger datasets show slight degradation, likely from underfitting.}
  \label{tab:datasetbench}
  \centering
  \begin{tabular}{llllll}
    \toprule
    & & \multicolumn{4}{c}{\textsc{Dataset}}                   \\
    \cmidrule(r){3-6}
    \textsc{Model}   &  \textsc{Method} & Cityscapes  & ACDC   & BDD100K & SHIFT  \\
    \midrule
    Baseline & - & 55.2 & 45.0 & \cellcolor[HTML]{FFEBDB}\textbf{57.2} & \cellcolor[HTML]{FFEBDB}\textbf{69.9}     \\
    Student-Teacher & Semi-Supervised & 55.4  & 48.2 & 56.2 & 68.6     \\
    Pseudo-Label & Label Transfer & 58.5  & 50.7 & 56.1 & 68.9    \\
    LAT & Label Transfer & \cellcolor[HTML]{FFEBDB}\textbf{60.0} & \cellcolor[HTML]{FFEBDB}\textbf{53.4} & 56.1 & 69.3     \\
    \bottomrule
  \end{tabular}
\end{table}

\paragraph{LAT improves performance across benchmark scenarios.}  
As shown in Table~\ref{tab:classbench}, LAT significantly outperforms the baseline in scenarios involving datasets with differing class granularity.  
It also yields substantial gains for smaller datasets when combined with much larger ones, as illustrated in Table~\ref{tab:datasetbench}.  
While we observe a slight performance drop for large datasets, this may be attributed to underfitting, which we analyse in a later section.

\paragraph{Domain gap limits student-teacher performance.} 
When compared to other models that use exponential moving average (EMA) without full semi-supervised training, student-teacher approaches consistently underperform, even falling below the baseline.  
One likely explanation is that domain gaps between datasets cause pseudo-label errors to propagate through the teacher model during training~\cite{li2022cross,kennerley2023tpcnet}.  
In contrast, LAT avoids this issue by retaining ground-truth labels during label transfer, which serve as reliable anchors for supervising pseudo-label refinement.

\begin{table}[ht]
  \caption{\textbf{Downstream AP} with extended training. Longer schedules help larger datasets recover, but gains remain smaller than for low-data regimes.}
  \label{tab:abalonger}
  \centering
  \begin{tabular}{lllll}
    \toprule
    & \multicolumn{4}{c}{Dataset}                   \\
    \cmidrule(r){2-5}
      & Cityscapes  & ACDC   & BDD100K & SHIFT  \\
    \midrule
    Baseline &  55.2 & 45.0 & 57.2 & 69.9     \\
    Strict &  60.0 & \cellcolor[HTML]{FFEBDB}\textbf{53.4} & 56.1 & 69.3     \\
    Double Iterations &  \cellcolor[HTML]{FFEBDB}\textbf{60.2}  & 53.3 & \cellcolor[HTML]{FFEBDB}\textbf{57.6} & \cellcolor[HTML]{FFEBDB}\textbf{70.9}    \\
    \bottomrule
  \end{tabular}
\end{table}

\paragraph{Extended training mitigates underfitting in large datasets.}
Larger datasets such as BDD100K and SHIFT may initially exhibit degraded performance when additional data is naively introduced, due to underfitting within the fixed training regime.  
However, extending the training schedule allows the downstream detector to better leverage the expanded supervision without compromising target alignment.  
As shown in Table~\ref{tab:abalonger}, performance on larger datasets improves with additional training iterations, while smaller datasets remain largely unaffected.

\begin{table}\bitsmall
    \parbox{.52\linewidth}{
   \caption{\textbf{Downstream AP} with different \textbf{clamping strategies} in SFF attention. Moderate scaling yields the best results.}
  \label{tab:abaclamp}
    \centering
  \begin{tabular}{lllll}
    \toprule
    & & \multicolumn{3}{c}{\textsc{Dataset}}                   \\
    \cmidrule(r){3-5}
       & Value & Cityscapes  & nuImages & Waymo  \\
    \midrule
    Clamping & $1/N$ & 57.2 & 40.9 & 47.2     \\
    Scaling & $1/N$ & 56.1 & 40.6 & 47.1     \\
    Clamping & $1/\sqrt{N}$ & 57.7 & 41.2 & 47.0    \\
    Scaling & $1/\sqrt{N}$ & \cellcolor[HTML]{FFEBDB}\textbf{59.6} & \cellcolor[HTML]{FFEBDB}\textbf{41.5}  & \cellcolor[HTML]{FFEBDB}\textbf{47.9}     \\
    \bottomrule
  \end{tabular}
  }
  \hfill
  \parbox{.45\linewidth}{
  \caption{\textbf{Downstream AP} under different \textbf{training strategies}. Fine-tuning after mixed training gives the highest performance.}
  \label{tab:abatrain}
  \centering
  \begin{tabular}{llll}
    \toprule
    & \multicolumn{3}{c}{\textsc{Dataset}}                   \\
    \cmidrule(r){2-4}
      Training  & Cityscapes  & nuImages & Waymo  \\
    \midrule\
    50/50 Batch & 57.5 & 40.5  & 46.8     \\
    Mixed Batch & 58.9 & 40.0  & 47.1    \\
    Fine-Tuning & \cellcolor[HTML]{FFEBDB}\textbf{59.6} & \cellcolor[HTML]{FFEBDB}\textbf{41.5}  & \cellcolor[HTML]{FFEBDB}\textbf{47.9}     \\
    \bottomrule
  \end{tabular}
  }
  \end{table}

\paragraph{Clamping strategy influences attention quality.}
Table~\ref{tab:abaclamp} presents our ablation of the clamping mechanism used in the SFF similarity matrix.  
We find that scaling at $1/\sqrt{N}$ consistently outperforms other clamping strategies, providing a balanced trade-off between preserving informative attention weights and suppressing noisy contributions.  
Stronger scaling values lead to a substantial drop in performance, suggesting that overly aggressive attenuation discards useful feature interactions during value matrix generation.  
Although clamping is a simple operation, it alters the relative weighting across features, which can negatively affect learning when not appropriately tuned.

\paragraph{Fine-tuning remains most effective.} 
We evaluate several strategies for training the downstream detector in Table~\ref{tab:abatrain}.  
These strategies vary the composition of source and target datasets within training batches: \textit{50/50 Batch} refers to batches containing an equal number of samples from each domain, while \textit{Mixed Batch} denotes random mixing of source and target samples.  
\textit{Fine-Tuning} modifies the Mixed Batch regime by replacing the final 10,000 training iterations with batches containing only source dataset samples.  
While all strategies outperform the baseline, we observe that pretraining on a mixed dataset followed by fine-tuning leads to the most consistent gains, highlighting the benefit of learning generaliable features before domain-specific adaptation.

\begin{table}[h!]
  \caption{\textbf{Downstream AP} on \textbf{Synscapes $\rightarrow$ Cityscapes}. LAT matches LGPL despite the simpler transfer setup.}
  \label{tab:syn}
  \centering
  \begin{tabular}{lll}
    \toprule
      Model & Def-DETR  & Faster RCNN  \\
    \midrule
    Baseline & 32.9 & 38.7     \\
    Pseudo-Label   & 30.7 & 36.9      \\
    Pseudo-Label + Filtering   & 33.0 & 39.1      \\
    LGPL \cite{liao2024transferring}   & 34.5 & \cellcolor[HTML]{FFEBDB}\textbf{39.7}      \\
    LAT   & \cellcolor[HTML]{FFEBDB}\textbf{34.7} & 39.6      \\
    \bottomrule
  \end{tabular}
\end{table}

\paragraph{Performance on simpler transfer protocols.}
We compare our method to LGPL~\cite{liao2024transferring} in Table~\ref{tab:syn}, using the synthetic-to-real transfer setting from Synscapes~\cite{synscapes} to Cityscapes~\cite{Cordts2016Cityscapes}.
We consider this a simpler transfer scenario, as both datasets share identical class labels and exhibit similar semantic structures.
Moreover, the setup involves a one-to-one label space transfer, reducing the need for LAT’s full design capabilities, such as many-to-one label alignment and the performance gains that emerge from integrating multiple source datasets.
Nevertheless, LAT matches the performance of the state-of-the-art LGPL method, demonstrating its effectiveness even under minimal transfer complexity.
We note that LGPL results are reported directly from the original paper using mAP@[.5:.95] as a metric, as public code was not available at the time of writing.

\section{Conclusion}

Label-Aligned Transfer (LAT) addresses the challenge of integrating object detection datasets with heterogeneous label spaces.
LAT modifies the standard two-stage detector by replacing the region proposal network with a \textit{Privileged Proposal Generator} (PPG), which incorporates both ground-truth and pseudo-label proposals from multiple source label spaces. 
A \textit{Semantic Feature Fusion} (SFF) module further refines region features by injecting privileged, class-aware context via attention over overlapping proposals. 
Our experiments demonstrate that LAT consistently improves target-domain performance, with gains of +4.2AP and +4.8AP on benchmarks evaluating high-to-low class granularity and small-to-large dataset transfer, respectively. 
LAT not only achieves state-of-the-art performance in complex multi-dataset settings, but also performs competitively under simpler one-to-one transfer protocols, highlighting its flexibility and generality.

\paragraph{Limitations.}
Our method assumes that annotations are required exclusively in a predefined target label space.  
Merging multiple label spaces or extending the target label space with additional classes is not currently supported.  
We aim to address this in future work by introducing mechanisms that provide greater flexibility in adapting or modifying the target label space.

\paragraph{Acknowledgements.} We would like to thank Dr. Wang Jian-Gang (A*STAR, I2R) for his support during this project.

\bibliography{unified}

\end{document}